\documentclass[runningheads]{llncs}
\usepackage{graphicx}
\usepackage{multirow}
\usepackage{float}
\usepackage{amsmath}
\usepackage{xcolor}
\usepackage{mwe}
\usepackage{hyperref}

\usepackage{subfig}
\usepackage{xr}
\usepackage{amssymb}
\usepackage{makecell}
 \usepackage{booktabs}

\def \loss {{\cal{L}}}

\def \latent{x}
\def \yb{\mathbf{y}}

\usepackage{xr-hyper}
\usepackage{hyperref}

\begin{document}
\title{Decoupled conditional contrastive learning with variable metadata for prostate lesion detection}
 \titlerunning{Decoupled conditional contrastive learning with variable metadata}
%
\author{Camille Ruppli \inst{1,3}
\and
Pietro Gori\inst{1} \and
Roberto Ardon\inst{3} \and
Isabelle Bloch \inst{2,1}
}

\authorrunning{C. Ruppli et al.}

\institute{
LTCI, Télécom Paris, Institut Polytechnique de Paris, Paris, France \and
Sorbonne Universit\'e, CNRS, LIP6, Paris, France \and
Incepto Medical, Paris, France
}

\maketitle             
\begin{abstract}

Early diagnosis of prostate cancer is crucial for efficient treatment. Multi-parametric Magnetic Resonance Images (mp-MRI) are widely used for lesion detection.  
The Prostate Imaging Reporting and Data System (PI-RADS) has standardized interpretation of prostate MRI by defining a score for lesion malignancy. PI-RADS data is readily available from radiology reports but is subject to high inter-reports variability.
We propose a new contrastive loss function that leverages weak metadata
with multiple annotators per sample and takes advantage of inter-reports variability by defining metadata confidence.
By combining metadata of varying confidence with unannotated data into a single conditional contrastive loss function, we report a 3\% AUC increase on lesion detection on the public PI-CAI challenge dataset.
 
 Code is available at : \url{https://github.com/camilleruppli/decoupled_ccl}
\keywords{Contrastive Learning  \and Semi-supervised Learning \and Prostate cancer segmentation}
\end{abstract}

\section{Introduction}

\noindent \textbf{Clinical context. } Prostate cancer is the second most common cancer in men worldwide. Its early detection is crucial for efficient treatment. Multi-parametric MRI has proved successful to increase diagnosis accuracy~\cite{Rouvire2019UseOP}. 
Recently, deep learning methods have been developed to automate prostate cancer detection~\cite{Bhattacharya2021SelectiveIA,Saha2021EndtoendPC,Yu2020DeepAP}.
Most of these methods rely on datasets of thousands of images, where lesions are usually manually annotated and classified by experts. 
This classification is based on the Prostate Imaging Reporting and Data System (PI-RADS) score, which ranges between 1 and 5, and associates a malignancy level to each lesion or to the whole exam (by considering the highest lesion score)~\cite{Turkbey2019ProstateIR}. 
This score is widely used by clinicians and is readily available from radiology reports. However, it is rather qualitative and subject to low inter-reader reproducibility~\cite{Smith2019IntraAI}. Images can also be classified using biopsy results, as in the PI-CAI dataset~\cite{PICAI_BIAS}. This kind of classification is usually considered more precise (hence often taken as ground truth), but is also
more costly to obtain and presents a bias since only patients with high PI-RADS scores undergo a biopsy.
Building a generic and automatic lesion detection method must therefore deal with the diversity of classification sources, radiology or biopsy, and the variability of classifications for a given exam.

\noindent \textbf{Methodological context. } In the past years, the amount of available medical imaging data has drastically increased. However, images are often either unannotated or weakly-annotated (\textit{e.g.,} a single PI-RADS score for the entire exam), as annotating each lesion is costly and time consuming. This means that usual supervised models cannot be used, since their performance highly depends on the amount of annotated data,  as shown in Table~\ref{tab:fully_sup}. 

\begin{table}[h!]
\centering
\caption{AUC at exam level (metrics defined in Section~\ref{sec:results}) on a hold out test set of the PI-CAI dataset of models trained from random initialization with 5-fold cross validation on a private dataset.}\label{tab:fully_sup}
\begin{tabular}{@{}cccccc@{}}
\toprule
 & 100\% \scriptsize($N_{train}$=1397) & 10\% \scriptsize($N_{train}$=139) & 1\% \scriptsize($N_{train}$=13) \\
\midrule
3D UNet & 0.80 (0.03) & 0.76 (0.02)  & 0.71 (0.04) \\
3D ResUnet & 0.79 (0.01) & 0.73 (0.02) & 0.64 (0.03) \\
\bottomrule
\end{tabular}
\end{table}

To take advantage of unannotated or weakly-annotated data during a pretraining step, self-supervised contrastive learning methods~\cite{pmlr-v119-chen20j,Grill2020BootstrapYO,He2019MomentumCF} have been developed.
Recent works have proposed to condition contrastive learning with class labels~\cite{Khosla2020SupervisedCL} or weak metadata~\cite{dufumier_integrating_2023,de_bruijne_contrastive_2021,Tsai2022ConditionalCL} to improve latent representations.
Lately, some works have also studied the robustness of supervised contrastive learning against noisy labels~\cite{xue_investigating_2022} proposing a new regularization~\cite{yi_learning_2022}.

 While contrastive pretraining has been widely applied to classification problems~\cite{pmlr-v119-chen20j,FernandezQuilez2022ContrastingAT,Grill2020BootstrapYO,He2019MomentumCF}, there have been few works about segmentation~\cite{basak_pseudo-label,Chaitanya2021LocalCL}.
 Recent works~\cite{Alonso2021SemiSupervisedSS,Chaitanya2021LocalCL,Zhao2020ContrastiveLF} propose to include pseudo labels at the pixel (decoder) level and not after the encoder, but, due to high computational burden, they can only consider 2D images and not whole 3D volumes.

Furthermore, many datasets contain several weak metadata at the exam level (e.g., PI-RADS score) obtained by different annotators. These weak metadata may have high inter- and intra-annotator variability, as for the PI-RADS score~\cite{Smith2019IntraAI}. This variability is rarely taken into account into self-supervised pre-training. Researchers usually either use all annotations, thus using the same sample several times, or they only use confident samples, based on the number of annotators and their experience or on the learned representations, as in~\cite{li_selective-supervised_2022}.

\noindent \textbf{Contributions. } Here, we aim to train a model that takes as input multi-parametric MRI exam and outputs a map where higher values account for higher lesion probability.
{\it Annotations} are provided by multiple annotators in the form of binary maps: segmentations of observed lesions. Only a small portion of the dataset  has annotations. A greater proportion has {\it metadata} information available from written  reports.
These metadata, referring to the whole exam, are either a (binary) biopsy grading (presence or absence of malignant lesion) or a PI-RADS score. For each exam, reports with a PI-RADS score are available from several radiologists, but the number of radiologists may differ among exams.

In the spirit of~\cite{Bovsnjak2023SemPPLPP,Dwibedi2021WithAL}, we propose to include confidence, measured as a degree of inter-reports variability on metadata, in a contrastive learning framework.
Our contributions are the following:
\begin{itemize}
    \item We propose a new contrastive loss function that leverages 
   weak metadata with multiple annotators per sample
   and takes advantage of inter-annotators variability by defining metadata confidence.
   \item We show that our method performs better than training from random initialization and previous pre-training methods on both the PI-CAI~\cite{saha_pi-cai_2022} public dataset and on a private multi-parametric prostate MRI dataset for prostate cancer lesion detection.
\end{itemize}

\section{Method}
We propose to apply contrastive pretraining to prostate lesion detection.
A lesion is considered detected if the overlap between the predicted lesion segmentation and the reference segmentation is above 0.1, as defined in~\cite{Saha2021EndtoendPC}. The predicted lesion masks are generated by a U-Net model~\cite{10.1007/978-3-319-24574-4_28} (since in our experiments this was the best model, see Table~\ref{tab:fully_sup}) fine-tuned after contrastive pretraining. 
In this section, we describe our contrastive learning framework defining confidence on metadata.

\subsection{Contrastive learning framework}\label{sec:method}
Contrastive learning (CL) methods train an encoder to bring close together latent representations of images of a positive pair while pushing further apart those of negative pairs. 
In unsupervised CL \cite{pmlr-v119-chen20j}, where no annotations or metadata are available, a positive pair is usually defined as two transformations of the same image and negative pairs as transformed versions of different images. Transformations are usually randomly chosen among a predefined family of transformations. The final estimated latent space is structured to learn invariances with respect to the applied transformations. 

In most CL methods, latent representations of different images are pushed apart \textit{uniformly}. The alignment/uniformity contrastive loss function proposed in~\cite{pmlr-v119-wang20k} is:
\begin{equation}
\label{eq:global_ua}
\begin{array}{cl}
    \loss_{NCE}  
    \displaystyle
    = \underbrace{\frac{1}{N}\sum_{i = 1}^N d_{ii}}_{\text{Global Alignment}} +
    \underbrace{\displaystyle
    \log\big( \frac{1}{N^2} \sum_{i,j = 1}^N e^{-d_{ij}} \big)}_{\text{Global Uniformity}}
\end{array}
\end{equation}
where $d_{ij}=||\latent_1^i - \latent_2^j||_2$, $\latent_1^i$ and $\latent_2^j$ are the encoder outputs of the transformed versions of images $i$ and $j$, respectively. 
However, as in many medical applications, our dataset contains discrete clinical features as metadata: PI-RADS scores and biopsy results per exam, that should be used to better define negative and positive samples.
To take metadata into account in contrastive pretraining, we follow the work of~\cite{dufumier_conditional_2021,de_bruijne_contrastive_2021}.
The authors introduce a kernel function on metadata~$y$ to \textit{condition} positive and negative pairs selection, defining the following loss function:
\begin{equation}
\label{eq:kernel_ua}
\begin{array}{cl}
    \loss_w  
    \displaystyle
    = \underbrace{\frac{1}{N}\sum_{i,j = 1}^N w (y_i,y_j) d_{ij}}_{\text{Conditional Alignment}} +
    \underbrace{\displaystyle
    \log\big( \frac{1}{N^2} \sum_{i,j = 1}^N (||w||_\infty-w (y_i,y_j)) e^{-d_{ij}} \big)}_{\text{Conditional Uniformity}}
\end{array}
\end{equation}
where $w$ is a kernel function measuring the degree of similarity between metadata~$y_i$ and $y_j$, $0 \leq w \leq 1$ and $||w||_\infty = w(y_i,y_i)=1$. The conditional alignment term brings close together, in the representation space, only samples that have a metadata similarity greater than 0, while the conditional uniformity term does not repel all samples uniformly but weights the repulsion based on metadata dissimilarity.
A schematic view of these two objective functions is shown in the supplementary material.

We apply this framework to metadata (PI-RADS scores and biopsy results) that can have high inter-report variability.

To simplify the problem and homogenize PI-RADS and biopsy scores, we decide to binarize both scores, following clinical practice and medical knowledge~\cite{Epstein2015The2I,Turkbey2019ProstateIR}. We set $y=0$ for PI-RADS 1 and 2, and $y=1$ for PI-RADS 4 and~5. We do not consider PI-RADS 3, since it has the highest inter-reader variability~\cite{Greer2019InterreaderVO} and low positive predictive value~\cite{Westphalen2020VariabilityOT}. This means that all exams with a PI-RADS 3 are considered deprived of metadata. For each exam $i$, a set of $y$ values is available, noted  $\yb_i$. The number of annotations may differ among subjects (see Equation~\eqref{eq:kernel_decoupled} for a definition of $w$ in such cases).
For a biopsy result (defining an ISUP classification~\cite{Epstein2015The2I}), we set $y=0$ if ISUP $\leq 1$ and $y=1$ if ISUP $\geq 2$.

To take advantage of the entire dataset, we also consider unannotated data for which metadata are not provided.
When computing the loss function on an exam without metadata (no $\yb$ associated),
we use the standard (unsupervised) contrastive loss function, as defined in~\cite{pmlr-v119-chen20j}.
This leads to the following contrastive loss function: 
\begin{equation}
\label{eq:kernel_ua_label_unlab}
    \begin{array}{ll}
    {\cal{L}}_w = &
    \left.
        \begin{array}{ll}
        \displaystyle
            \frac{1}{|A|}\sum_{i \in A} \Big( \sum_{j \in A} w (\yb_j,\yb_i) ||x_1^i - x_2^j|| \Big)  \\
         + 
         \displaystyle
        \log\big(\frac{1}{|A|^2}\sum_{i,j \in A} (1 - w(\yb_i,\yb_j)) e^{-||x_1^i - x_2^j||}\big) \\
        \end{array}
     \hphantom{xxxxxxx}\right \}\text{\tiny with metadata} \\
&  \left.
        \begin{array}{ll}
        \displaystyle
        + \frac{1}{|U|}\sum_{i \in U} \Big( ||x_1^i - x_2^i||\Big) 
        +  
        \displaystyle
        \log \Big( \frac{1}{|U|^2} \sum_{\substack{i,j \in U \\ i \neq j}} e^{-||x_1^i - x_2^j||} \Big)
        \end{array}
\right \}\text{\tiny without metadata}
    \end{array}
\end{equation} 
where $A$ (resp. $U$) is the subset with (resp. without) associated $\yb$ metadata.

Since the number of annotations may be different between two subjects $i$ and $j$, we cannot use a standard kernel, as the RBF in~\cite{de_bruijne_contrastive_2021}. We would like to take into account metadata confidence, namely agreement among annotators. In the following, we propose a new kernel $w$ that takes metadata confidence into account.

\begin{figure}[H]\centering
\includegraphics[width=\linewidth]{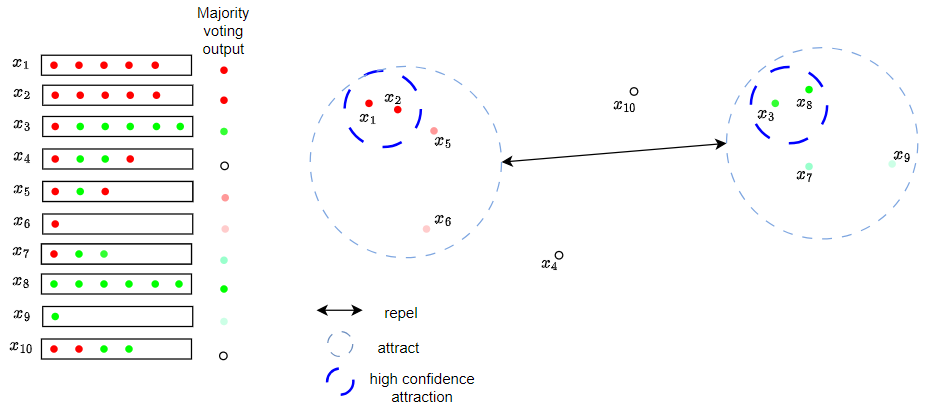}
\caption{Given a set of exams $x_{i \in [1,10]}$, $\yb_i$ is represented as a list of colored points. Confidence ($c$) is represented with color saturation: darker means more confident. Exams such that $c(\yb_i) = 0$ (no decision from majority voting) are considered as unlabeled and uncolored. Exams such that $c(\yb_{i,j}) = 1$ and $y_i = y_j$, e.g. $(x_1, x_2)$ (resp. $(x_3, x_8)$), will be strongly attracted while less attracted to patients with $c(\yb_i) < 1$, e.g. $x_{5,6}$ (resp. $x_{7,9}$). Groups of exams with different $y$ scores are repelled.} \label{fig:confidence_loss_schema}
\end{figure}

\smallskip
\noindent \textbf{Confidence. }
Our measure of confidence is based on the discrepancy between the elements of vector $\yb$ and their most common value (or majority voting). For exam $i$, if $y_i$ is the most common value in its metadata vector $\yb_i$  = $[y_{i0}, y_{i1}, ... y_{in-1}]$ with $n$ the number of available scores, confidence $c$ is defined as:
\begin{equation}
\label{eq:confidence}
c(\yb_i) = 
\left\{
\begin{array}{lcl}
\epsilon &\text{if }& n=1 \\
\\
2 \times \left(\dfrac{\sum_{k=0}^{n-1} \delta(y_{ik}, y_i)}{n}-\frac{1}{2}\right) &\text{if }& n>1
\end{array}
\right.
\end{equation}
where $\delta$ is the Dirac function and $\epsilon = 0.1$\footnote[1]{ The maximum number of metadata available for an exam is $n=7$, the minimal achievable confidence value is thus $c = 2(4/7 -1/2) > 0.14$. We fix $\epsilon$ so that the confidence for $n=1$ is higher than $0$ but less that the minimal confidence when $n$ is odd.}. 
$c(\yb_i) \in [0, 1]$, $0$ is found when an even number of opposite scores is obtained and the majority voting cannot provide a decision. In that case the associated exam will be considered as deprived of metadata. The proposed kernel then reads :
\begin{equation}
\label{eq:kernel_decoupled}
  w(\yb_i, \yb_j) =
  \setlength{\arraycolsep}{0pt}
  \renewcommand{\arraystretch}{1.2}
  \left\{\begin{array}{l @{\quad} l r}
        1           & \text{if } i = j & \makecell{\text{\scriptsize(exam against its} \\ \text{\scriptsize own transformed version)}} \\        
        c_{ij} & \text{if } y_i = y_j  \text{ and } i \neq j &               \makecell{\text{\scriptsize (different exams,} \\ \text{\scriptsize same majority voting)}} \\        
        0   & \text{if } y_i \neq y_j  \text{ and } i \neq j & \makecell{\text{\scriptsize (different exams,} \\ \text{\scriptsize different majority voting)}} \\
  \end{array}\right.
\end{equation}
 where $c_{ij} = \min(c(\yb_i), c(\yb_j))$. For two given exams $i$ and $j$, the proposed model is interpreted as follows:

\begin{itemize}
    \item If both metadata confidences are maximal ($c_{ij} = 1$), $w(\yb_i, \yb_j)$ will be equal to 1 and full alignment will be computed.
    \item If either metadata confidence is less than 1, $w(\yb_i, \yb_j)$ value will be smaller and exams will not be fully aligned in the latent space. The less confidence, the less aligned exams $i$ and $j$ representations will be.
    \item If confidence drops to zero for either exam, the exam will only be aligned with its own transformed version.
\end{itemize}
Similarly to decoupled CL~\cite{5c6b8c6afdc64b258cae1ad4e4d025c2}, we design $w$ such that the second term of Equation~\eqref{eq:kernel_ua_label_unlab} does not repel samples with identical metadata most common value and maximal confidence ($c_{ij} = 1$). See Figure~\ref{fig:confidence_loss_schema} for a schematic view. 

\subsection{Experimental settings}

\textbf{Datasets. } Experiments were performed on a private dataset of 2415 multi parametric MRI prostate exams among which 1397 have annotations (metadata and manual lesion segmentation) provided by multiple radiologists (up to 7). 
We also use the public PI-CAI dataset~\cite{PICAI_BIAS} composed of 1500 exams and 1295 annotations. 
In all learning steps, we used T2 weighted (T2w), apparent diffusion coefficient (ADC) and diffusion weighted (with the highest available b-value in the exam) sequences. As in~\cite{Saha2021EndtoendPC,Yu2020FalsePR}, we use the prostate ternary segmentation (background, peripheral zone and central zone), generated from an independent process on T2w sequences. We thus learn from a total of four volumes considered as registered. 
Pretraining is performed on data from both datasets on 3915 exams. Fine-tuning is performed with 1\% and 10\% of these exams using cross validation (see \textbf{Implementation details}).

\smallskip

\noindent\textbf{Implementation details. } We pretrain the chosen U-Net encoder followed by a projection head. Similarly to nn U-Net~\cite{Isensee2020nnUNetAS}, the encoder is a fully convolutional network where spatial anisotropy is used (e.g. axial axis is downsampled with a lower frequency since MRI volumes often have lower resolution in this direction). It is composed of four convolution blocks with one convolution layer in each block and takes the four sequences as input in channel dimension. The projection head is a two-layer perceptron as in~\cite{pmlr-v119-chen20j}. We train with a batch size of 16 for 100 epochs and use a learning rate of $10^{-4}$. Following the work of~\cite{FernandezQuilez2022ContrastingAT} on contrastive learning for prostate cancer triage, we use a random sampling of rotation, translation and horizontal flip to generate the transformed versions of the images.

To evaluate the impact of contrastive pretraining at low data regime, we perform fine-tuning with 10\% and 1\% of annotated exams.
The contrastive pretrained encoder is used to initialize the U-Net encoder, the whole encoder-decoder architecture is then fine-tuned on the supervised task.
Fine-tuning is performed with \textbf{5-fold cross validation} with both datasets using the pretrained encoder. Using 1\% (resp. 10\%) of annotated data, each fold has 39 (resp. 269) training data and 12 (resp. 83) validation data. We build a hold out test set of 500 volumes\footnote[2]{ The 100 validation cases on the PI-CAI challenge website being hidden we could not compare our methods to the leaderboard performances.}, not used during any training step with data from both datasets to report our results.
We also compared fine-tuning from our pretrained encoder to a model trained from random initialization. Fine-tuning results with 10\% of
annotated data are reported the in supplementary material.

\smallskip

\noindent\textbf{Computing infrastructure. } Optimizations were run on GPU NVIDIA T4 cards.

\section{Results and discussion}\label{sec:results}
The 3D U-Net network
outputs lesions segmentation masks which are thresholded, following the dynamic thresholding proposed in~\cite{Bosma2021AnnotationefficientCD}, and of which connected components are computed. For each connected component, a detection probability is assigned as the maximum value of the network output in this component. The output of this post-processing is a binary mask associated with a detection probability per lesion. We compute the overlap between each lesion mask and the reference mask. A lesion is considered as a true positive (detection) if the overlap with the reference is above 0.1 as defined in~\cite{Saha2021EndtoendPC}. This threshold is chosen to keep a maximum number of lesions to be analyzed for AUC computation. Different thresholds values are then applied for AUC computation.

As in~\cite{Saha2021EndtoendPC,Yu2020DeepAP}, lesion detection probability is used to compute AUC values at exam and lesion levels, and average precision (mAP). To compute AUC at exam level we take, as ground truth, the absence or presence of a lesion mask, and, as a detection probability, the maximum probability of the set of detected lesions. At lesion level, all detection probabilities are considered and thresholded with different values, which amounts to limiting the number of predicted lesions. The higher this threshold, the lower are sensitivity and the number of predicted lesions, and the higher is specificity.

Results are presented in Table~\ref{results}. For both datasets, we see that including metadata confidence to condition alignment and uniformity in contrastive pretraining yields better performances than previous state of the art approaches and random initialization. 
The discrepancy between PI-CAI and private mAP is due to the nature of the dataset: the PI-CAI challenge was designed to detect lesions confirmed by biopsy, while our private dataset contains lesions not necessarily confirmed by biopsy. Our private dataset contains manually segmented lesions that might be discarded if biopsy was performed. The model being fine-tuned on both datasets, PI-CAI exams are overly segmented which leads to lower mAP values (since our model tends to over-segment on biopsy ground truths). 
For our clinical application, which aims to reproduce radiologist responses, this is acceptable. We report significant performance improvement at very low data regime (1\% annotated data) compared to existing methods which is a framework often encountered in clinical practice.

To assess the impact of our approach we perform different ablation studies (shown in the second part of Table~\ref{results}).

\noindent \textbf{High confidence} (HC row in Table~\ref{results}). For pretraining, only exams with confidence equal to 1 are considered but are not perfectly aligned ($c_{ij} = 0.8\delta(c_{ij},1)$ in Equation~\eqref{eq:kernel_decoupled}). We can see that considering only confident samples to condition contrastive learning decreased performances.

\noindent \textbf{Majority Voting} (Majority voting row in Table~\ref{results}). We removed confidence and used majority voting output for kernel computation. If two different exams have the same majority vote we set : $w(y_i, y_j) = 0.8$ in Equation~\eqref{eq:kernel_decoupled}, other $w$ values are kept unchanged. We can see that using majority voting output without taking confidence into account leads to decreased performances.

\noindent \textbf{Biopsy} (Biopsy row in Table~\ref{results}). We set the confidence of PI-CAI exams to 1 (increasing biopsy confidence) which amounts to setting $\epsilon = 1$ for PI-CAI exams in Equation~\eqref{eq:confidence}. No particular improvement is observed with this approach. 

\noindent \textbf{Global uniformity.} We remove the conditioning on uniformity. Exams are uniformly repelled rather than conditioning on metadata similarity for repulsion (which amounts to setting $w(\yb_i,\yb_j) = 0$ for the second term of Equation~\eqref{eq:kernel_ua_label_unlab}). Removing uniformity conditioning yields lower performances than 
the proposed approach (GlU row in Table~\ref{results}).

\begin{figure}\centering
\includegraphics[scale=0.4]{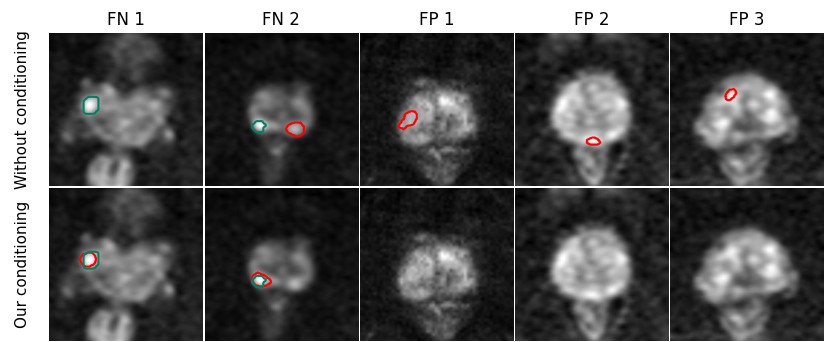}
\caption{Examples of false negative (FN) and false positive (FP) cases (first row) corrected by the proposed method (second row). Reference segmentation: green overlay, predicted lesions: red overlay} \label{fig:qualitative_results}
\end{figure}

Figure~\ref{fig:qualitative_results} shows the impact of our pretraining method on the finetuned U-Net outputs. Without conditioning, some lesions are missed (cases FN 1, FN 2) and others are falsely detected (cases FP 1, 2 and 3). Adding the conditioned pretraining removed these errors. More examples are provided in the supplementary material.

\begin{table}
\centering
\caption{5-fold cross validation mean AUC and mAP after fine-tuning on PI-CAI and private datasets with 1\% of annotated data (standard deviation in parentheses)}\label{results}
\small
\begin{tabular}{@{}ccccccccc@{}}
\toprule
Method & \multicolumn{2}{c}{AUC exam} & \phantom{.} & \multicolumn{2}{c}{AUC lesion} & \phantom{.} & \multicolumn{2}{c}{mAP} \\
\cmidrule{2-3} \cmidrule{5-6}  \cmidrule{8-9}
& PI-CAI & Private && PI-CAI & Private && PI-CAI & Private \\ \midrule
Random init & 0.68 \scriptsize(0.06) & 0.74 \scriptsize(0.03) &&	0.73 \scriptsize(0.11) & 	0.70 \scriptsize(0.05)	&& 0.27 \scriptsize(0.05) & 0.62 \scriptsize(0.03)  \\
Unif Align~\cite{pmlr-v119-wang20k} & 0.66 \scriptsize(0.07) & 0.72 \scriptsize(0.01)	&& 0.64 \scriptsize(0.13) & 0.68 \scriptsize(0.03)	&&	0.28 \scriptsize(0.07)  & 0.63 \scriptsize(0.03) \\
simCLR~\cite{pmlr-v119-chen20j} & 0.64 \scriptsize(0.07) & 0.73 \scriptsize(0.05) && 0.65 \scriptsize(0.08) & 0.68 \scriptsize(0.05) &&	0.22 \scriptsize(0.07) & 0.60 \scriptsize(0.06) \\
MoCo~\cite{He2019MomentumCF}  & 0.63 \scriptsize(0.08) & 0.71 \scriptsize(0.04)	&&	0.59 \scriptsize(0.12) & 0.64 \scriptsize(0.07)	&&	0.24 \scriptsize(0.10) & 0.58 \scriptsize(0.06) \\
BYOL~\cite{Grill2020BootstrapYO}  & 0.67 \scriptsize(0.06) & 0.72 \scriptsize(0.04) &&	0.66
\scriptsize(0.16) & 	0.68 \scriptsize(0.04)	 &&	0.26 \scriptsize(0.05) & 0.59 \scriptsize(0.04) \\
nnCLR~\cite{Dwibedi2021WithAL} & 0.57 \scriptsize(0.08) & 0.73 \scriptsize(0.05) &&	0.49 \scriptsize(0.09) & 	0.62 \scriptsize(0.06) &&	0.21 \scriptsize(0.05) & 0.59 \scriptsize(0.05) \\
Ours & \textbf{0.70} \scriptsize(0.05) & \textbf{0.75} \scriptsize(0.03) &&	\textbf{0.75} \scriptsize(0.10) & 0.71 \scriptsize(0.03) &&	\textbf{0.30} \scriptsize(0.09) & \textbf{0.63} \scriptsize(0.04)\\
\hline
 GlU  & 0.60 \scriptsize(0.05) & 0.74 \scriptsize(0.03)	&&	0.60 \scriptsize(0.12) & \textbf{0.73} \scriptsize(0.02) &&	0.23 \scriptsize(0.05) & 0.63 \scriptsize(0.04) \\
Biopsy & 0.64 \scriptsize(0.06) & 0.73 \scriptsize(0.03) && 0.69 \scriptsize(0.06)& 	0.70 \scriptsize(0.03) && 0.24 \scriptsize(0.05) & 0.62 \scriptsize(0.04) \\
HC & 0.66 \scriptsize(0.08) & 0.75 \scriptsize(0.04) && 0.60 \scriptsize(0.06) & 0.67 \scriptsize(0.03) && 0.28 \scriptsize(0.09) & 0.62 \scriptsize(0.03) \\
Majority voting & 0.63 \scriptsize(0.06) & 0.74 \scriptsize(0.02) && 0.62 \scriptsize(0.06) & 0.69 \scriptsize(0.04) && 0.28 \scriptsize(0.07) & 0.61 \scriptsize(0.04) \\
\bottomrule
\end{tabular}
\end{table}

\section{Conclusion}
We presented a new method to take the confidence of metadata, namely the agreement among annotators, into account in a contrastive pretraining. We proposed a definition of metadata confidence and a new kernel to condition positive and negative sampling.
The proposed method yielded better results for prostate lesion detection than existing contrastive learning approaches on two datasets. 


\bibliographystyle{splncs04}
\bibliography{ref.bib}

\end{document}